\documentclass[runningheads]{llncs}
\usepackage{subcaption}
\usepackage{xspace}
\usepackage{graphicx}
\usepackage{comment}
\usepackage{amsmath,amssymb} 
\usepackage{color}
\usepackage{ifthen}

\definecolor{green}{rgb}{0.3, 0.5, 0.0}
\newcommand{\smpl}{SMPL\xspace}
\newcommand{\supmat}{Sup.~Mat.\xspace}
\newcommand{\modelname}{STAR\xspace}

\begin{document}

\title{STAR: Sparse Trained Articulated Human Body Regressor}

\titlerunning{STAR}

\author{Ahmed A. A. Osman  \and
Timo Bolkart \and Michael J. Black}

\authorrunning{Osman et al.}
\institute{Max Planck Institute for Intelligent Systems, T{\"u}bingen, Germany  \\
\email{\{aosman,tbolkart,black\}@tuebingen.mpg.de}}
\maketitle

\begin{abstract}
The \smpl body model is widely used for the estimation, synthesis, and analysis of 3D human pose and shape. 
While popular, we show that SMPL has several limitations and introduce \modelname, which is quantitatively and qualitatively superior to SMPL.
First, \smpl has a huge number of parameters resulting from its use of global blend shapes.
These dense pose-corrective offsets relate every vertex on the mesh to all the joints in the kinematic tree,  capturing spurious long-range correlations.
To address this, we define per-joint pose correctives and learn the subset of mesh vertices that are influenced by each joint movement. 
This sparse formulation results in more realistic deformations and significantly reduces the number of model parameters to 20\% of \smpl. 
When trained on the same data as \smpl, \modelname generalizes better despite having many fewer parameters.
Second, SMPL factors pose-dependent deformations from body shape while, in reality, people with different shapes deform differently.
Consequently, we learn shape-dependent pose-corrective blend shapes that depend on both body pose and BMI.
Third, we show that the shape space of SMPL is not rich enough to capture the variation in the human population.
We address this by training \modelname with an additional 10,000 scans of male and female subjects, and show that this results in better model generalization. 
\modelname is compact, generalizes better to new bodies and is a drop-in replacement for SMPL. 
STAR is publicly available for research purposes at \url{http://star.is.tue.mpg.de}.
\end{abstract}
\section{Introduction}
\begin{figure}[t]
	\centering
\includegraphics[width=\textwidth]{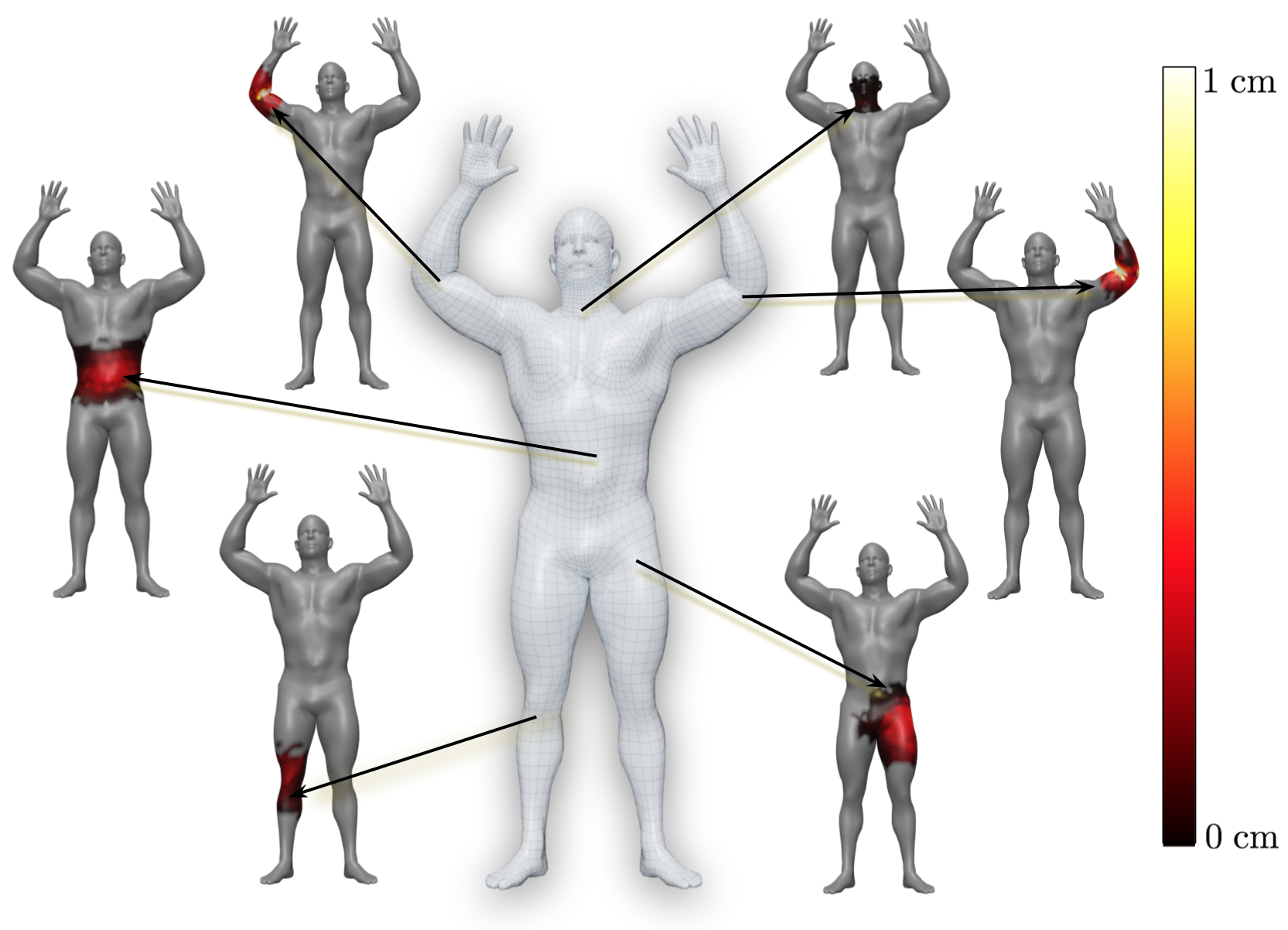}
\caption{\textbf{Sparse Local Pose Correctives:} STAR factorizes
  pose-dependent deformation into a set of sparse and spatially local
  pose-corrective blendshape functions, where each joint influences
  only a sparse subset of mesh vertices. The white mesh is
  \modelname fit to a 3D scan of a professional body builder. 
The arrows point to joints in the \modelname kinematic tree and the
corresponding predicted corrective offset for the joint. The heat map
encodes the magnitude of the corrective offsets. The joints have no
influence on the gray mesh vertices.}
\label{fig:sparse_decomposition}
\end{figure}

Human body models are widely used to reason about 3D body pose and shape in images and videos. 
While several models have been proposed \cite{Allen:2003:SHB:882262.882311,Anguelov05,chen2013tensor,Freifeld:ECCV:2012,hasler2009statistical,Hirshberg:ECCV:2012,Pons-Moll:2015:DMD:2809654.2766993,Seo:2003,xu2020ghum}, 
\smpl \cite{SMPL:2015} is currently the most widely use in academia and industry.
\smpl is trained from thousands of 3D scans of people and captures the statistics of human body shape and pose.
Key to \smpl's success is its compact and intuitive parametrization, decomposing the 3D body into pose parameters $\vec{\theta} \in \mathbb{R}^{72}$  corresponding to axis angle rotations of $24$ joints and shape $\vec{\beta} \in \mathbb{R}^{10}$ capturing subject identity (the number of shape parameters can be as high as 300 but most research uses only 10).
This makes it useful to reason about 3D human body pose and shape given sparse measurements, such as IMU accelerations \cite{MuVS:3DV:2017,DIP:SIGGRAPHAsia:2018,SIP2017}, sparse mocap markers \cite{Loper:SIGASIA:2014,amass2019} or 2D key points in images and videos \cite{alldieck2018detailed,kanazawa2018end,kocabas2019vibe,SPIN:ICCV:2019,pavlakos2018learning,Ruegg:AAAI:2020,tan2017indirect,zanfir2018monocular}.

\begin{figure}[t]
\begin{subfigure}{0.5\textwidth}
	\centering
\includegraphics[width=0.8\textwidth]{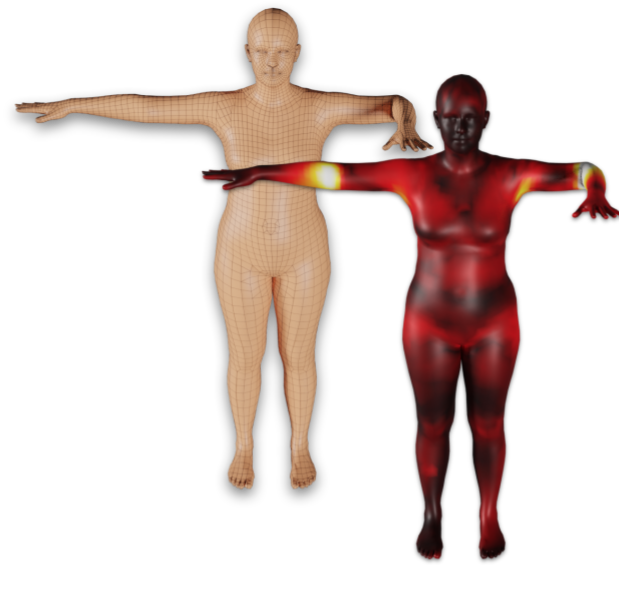}
	\caption{}
\label{fig:spur_corr}
\end{subfigure}
\begin{subfigure}{0.5\textwidth}
		\centering
	\includegraphics[width=0.8\textwidth]{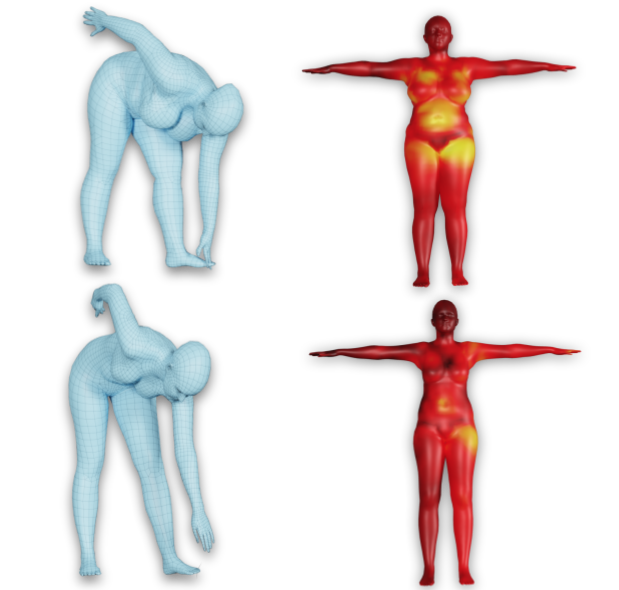}
	\caption{}
		\label{fig:shape_deform}
\end{subfigure}

\caption{\textbf{SMPL Limitations}: Examples of some \smpl limitations. Heat maps illustrate the magnitude of the pose-corrective offsets. Fig.~\ref{fig:spur_corr} highlights the spurious long-range correlations learned by the \smpl pose corrective blend shapes. Bending one elbow results in a visible bulge in the other elbow. Fig.~\ref{fig:shape_deform} shows two subjects registrations (show in blue) with two different body shapes (High BMI) and (Low BMI). While both are in the same pose, the corrective offsets are different since body deformation are influenced by both body pose and body shape.  The  \smpl pose corrective offsets are the same regardless of body shape.}
\label{fig:fig_smpl_problem_teaser}
\end{figure}

While \smpl is widely used it suffers from several drawbacks. 
\smpl augments traditional linear blend skinning (LBS) with pose-dependent corrective offsets that are learned from 3D scans.
Specifically, \smpl uses a pose-corrective blendshape function $\mathcal{P}(\vec{\theta}): \mathbb{R}^{|\vec{\theta}|} \rightarrow \mathbb{R}^{3N}$, where $N$ is the number of mesh vertices. 
The function $\mathcal{P}$ predicts corrective offsets for every mesh vertex such that, when the model is posed, the output mesh looks realistic. 
The function $\mathcal{P}$ can be viewed as a fully connected layer (FC), that relates the corrective offsets of every mesh vertex to the elements of the part rotation matrices of all the body joints. 
This dense blendshape formulation has several drawbacks. 
First, it significantly inflates the number of model parameters to $> 4.2$ million, making \smpl prone to overfitting during training. 
Even with numerous regularization terms, the model learns spurious correlations in the training set, as shown in  Figure \ref{fig:spur_corr};  moving one elbow causes a bulge in the other elbow.

This is problematic for graphics, model fitting, and deep learning. 
The dense formulation causes dense spurious gradients to be propagated through the model. 
A loss on the mesh surface back propagates spurious gradients to geodesically distant joints. 
The existing formulation of the pose corrective blend shapes limits the model compactness and visual realism.

To address this, we create a new compact human body model, called {\bf STAR} (Sparse Trained Articulated Regressor), that is more accurate than \smpl yet has sparse and spatially local blend shapes, such that a joint only influences a sparse set of vertices that are geodesically close to it. 
The original \smpl paper acknowledges the problem and proposes a  model called SMPL-LBS-Sparse that restricts the pose corrective blend shapes such that a vertex is only influenced by joints with the highest skinning weights.
SMPL-LBS-Sparse, however, is less accurate than \smpl.

Our key insight is that  the influence of a body joint should be inferred from the training data. 
The main challenge is formalizing a model and training objective such that we learn meaningful joint support regions that are sparse and spatially local as shown in Figure \ref{fig:sparse_decomposition}. 
To this end we formalize a differentiable thresholding function based on the Rectified Linear Unit operator, \textbf{ReLU}, that learns to predict 0 activations for irrelevant vertices in the model.
The output activations are used to mask the output of the joint blendshape regressor to only influence vertices with non-zero activations. This results in a sparse model of pose-dependent deformation.

We go further in improving the model compactness. \smpl uses a Rodrigues representation of the joints angles and has a separate pose-corrective regressor for each element of the matrix, resulting in 9 regressors per joint. We switch to a quaternion representation with only 4 numbers per joint, with no loss in performance.
This, in combination with the sparsity, means that \modelname has  20\% of the parameters of SMPL.
We evaluate \modelname by training it on different datasets.
When we train \modelname on the same data as \smpl, we find that it is more accurate on held-out test data.
Note that the use of quaternions is an internal representation change from \smpl and transparent to users who can continue to use the \smpl pose parameters.

\smpl disentangles shape due to identity from shape due to pose.  This is a strength because it results in a simple model with additive shape functions.
It is also a weakness, however, because it cannot capture correlations between body shape and how soft tissue deforms with pose.
To address this we extend the existing pose corrective formulation by regressing the correctives using both body pose $\vec{\theta}$ and body shape $\vec{\beta}$. 
Here we use the second principal component of the of the body shape space, which correlates highly with Body Mass Index (BMI). 
This change results in more realistic pose-based deformations. 

\smpl is used in many fields such as apparel and healthcare because it captures the statistics of human body shape.
The \smpl shape space was trained using the CAESAR database, which contains 1700 male and 2107 female subjects. 
CAESAR bodies, however, are distributed according to the US population in 1990 \cite{CAESAR} and do not reflect global body shape statistics today.
Additionally, CAESAR's capture protocol dressed all women in the same sports-bra-type top, resulting in a female chest shape that does not reflect the diversity of shapes found in real applications.
We show that \smpl trained on CAESAR is not able to capture the variation in the more recent, and more diverse, SizeUSA dataset of 10,000 subjects (2845 male and 6436 female) \cite{SizeUSA}, and vice versa.
To address these problems, we train \modelname from the combination of CAESAR and SizeUSA scans and show that the complementary information contained in both datasets enables \modelname to generalize better to unseen body shapes.

We summarize our contributions by organizing them around impact areas where \smpl is currently used:
\begin{enumerate}
\item {\bf Computer vision:}
We propose a  compact model that is 80\% smaller than \smpl.
We achieve compactness in two ways: First, we formalize sparse corrective blend shapes and learn the set of vertices influenced by each joint. Second, we use quaternion features for offset regression. While \modelname is more compact than \smpl, it generalizes better on held-out test data. 
\item {\bf Graphics:}  Non-local deformations make animation difficult because changing the pose of one body part  affects other parts.  Our local model fixes this problem with \smpl.
\item {\bf Health:}  Realistic avatars are important in health research. We increase realism by conditioning the pose corrective blend shapes on body shape.  Bodies with different BMI produce different pose corrective blend shapes.  
\item {\bf Clothing Industry:}  Accurate body shape matters for clothing. 
We use the largest training set to date to learn body shape and show that previous models were insufficient to capture the diversity of human shape.
\end{enumerate}
The model is a drop-in replacement for \smpl, with the same pose and shape parametrization. 
We make the model with a 300 principal component shape space publicly available for research purposes at \url{http://star.is.tue.mpg.de}.
\section{Related Work}
\label{sec:related_work}

There is a long literature on 3D modelling of the human body, either manually or using data-driven methods.
We review the most related literature here with a focus on methods that learn bodies from data, pioneered by  \cite{Allen:TOG:2002,Anguelov05}.

\paragraph{Linear Blend Skinning.} 
Linear Blend Skinning (LBS), also known as Skeletal-Subspace Deformation (SSD) \cite{magnenat1988joint,magnenat1990human}, is the foundation for many existing body models because of its simplicity. 
With LBS, the mesh is rigged with an underling set of joints forming a kinematic tree where each mesh vertex $v_i$ is associated with $n$ body joints and corresponding skinning weights $w_i$. 
The transformations applied to each mesh vertex are a weighted function of the transformations of the associated $n$ joints. 
The skinning weights are typically defined by an artist or learned from data. 
In \smpl \cite{SMPL:2015} the skinning weights are initialized by an artist and fine tuned as part of the training process. 
Numerous works attempt to predict the skinning weights for arbitrary body meshes, e.g.~\cite{jacobson2012fast,liu2019neuroskinning}.

\paragraph{Pose Corrective Blend Shapes.} 
Although LBS is widely used, it suffers from well known shortcomings, which several method have been proposed to address.  Lewis \cite{Lewis:2000:PSD:344779.344862} introduces the pose space deformation model (PSD) where LBS is complemented with  corrective deformations. The deformations are in the form of  corrective offsets added to the mesh vertices  posed with LBS. The corrective deformations are related to the underlying kinematic tree pose.  
Weighted pose deformation (WPD) \cite{kurihara2004modeling,rhee2006real} adds pose corrective offsets to the base template mesh in the canonical (rest) pose before posing it with LBS, such that final posed mesh is plausible. 
Typically, such correctives are artist defined in key poses.  
Given a new pose, a weighted combination of correctives from nearby key poses is applied.
Allen et al.~\cite{Allen:TOG:2002} are the first to learn such corrective offsets from 3D scans of human bodies. 

\paragraph{Learned Models.}
The release of the CAESAR dataset of 3D scans \cite{CAESAR} enabled researchers to begin training statistical models of body shape  \cite{Allen:2003:SHB:882262.882311,Seo:2003}.
SCAPE \cite{Anguelov05} is the first model to learn a factored representation of body shape and pose. 
SCAPE models body deformations due pose and shape as triangle deformations and has been extended in many ways
\cite{chen2013tensor,Freifeld:ECCV:2012,hasler2009statistical,Hirshberg:ECCV:2012,PISHCHULIN2017276,Pons-Moll:2015:DMD:2809654.2766993}. SCAPE has several downsides, however. It requires a least-squares solver to create a valid mesh, has no explicit joints or skeletal structure, may not maintain limb lengths when posed, and is not compatible with graphics pipelines and game engines.

To address these issues, Loper et al.~\cite{SMPL:2015} introduced \smpl, which uses vertex-based corrective offsets. 
Like SCAPE, \smpl factors the body into shape dependent deformations and pose dependent deformations. 
\smpl is more accurate than SCAPE when trained on the same data and is based on LBS, making it easier to use.
\smpl is also the first model trained using the full CAESAR dataset \cite{CAESAR}, giving it a realistic shape space; previous methods used a subset of CAESAR or even smaller datasets.

\smpl models pose correctives as a linear function of the elements of the part rotation matrices.  
This results in 207 pose blend shapes with each one having a global effect.
Instead, we train a non-linear model that is linear in the pose (for good generalization) but non-linear in the spatial extent (to make it local). 
We adopt a unit quaternion representation and reduce that number of blend shapes from 207 to 23. 
These functions are not based on a single joint but rather on groups of joints, giving more expressive power.
We train the correctives using a non-linear function that encourages spatial sparsity in the blend shapes. 
This results in a model that is 80\% smaller than \smpl and reduces long-range spurious deformations.
Loper et al.~\cite{SMPL:2015} also proposed a sparse version of \smpl but found that it reduced accuracy.
In contrast, when trained on the same data, \modelname is more accurate than \smpl.
Additionally, we show that CAESAR is not sufficient and we train on more body shape data (14,000 scans in total) than any previous model.

SMPL and SCAPE factor body shape and pose-dependent shape change, but ignore correlations between them.
Several methods model this with a tensor representation \cite{chen2013tensor,hasler2009statistical}.
This allows them to vary muscle deformation with pose depending on the muscularity of the subject.
Here we achieve similar effects while keeping the benefits of simple models like \smpl.

\paragraph{Sparse Pose Corrective Blend Shapes.}
Human pose deformations are largely local in nature and, hence, the pose corrective deformations should be similarly local. 
Kry et al.~\cite{kry2002eigenskin} introduce EigenSkin to learn a localized model of pose deformations. 
\modelname is similar to EigenSkin in that it models localized joint support but, unlike EigenSkin we infer the joint support region from posed scan data without requiring a dedicated routine of manually posing joints. 
Neumann et al.~\cite{neumann2013sparse}, use sparse PCA to learn local and sparse deformations of  pose-dependent body deformations but do not learn a function mapping body pose to these deformations. 
In contrast, \modelname  learns sparse and local pose deformations that are regressed directly from the body pose. 
Contemporaneous with our work, GHUM \cite{xu2020ghum} builds on SMPL and its Rodrigues pose representation
but reduces the pose parameters (including face and hands) to a 32-dimensional latent code.
Pose correctives are linearly regressed from this latent representation with L1 sparsity, giving sparse correctives.

\section{Model}
\modelname is a vertex-based LBS model complemented with a learned set  of shape and pose corrective functions. Similar to \smpl, we factor the body shape into  the subject's intrinsic shape and pose-dependent deformations. In \modelname  we define a pose corrective function for each joint, $j$, in the kinematic tree. In contrast to \smpl, we condition the pose corrective deformation function on both body pose $\vec{\theta} \in \mathbb{R}^{|\vec{\theta}|}$ and shape $\vec{\beta} \in \mathbb{R}^{|\vec{\beta}|}$. Additionally,  during training, we use a non-linear activation function, $\phi(.)$, that selects the subset of mesh vertices relevant to the joint $j$. The pose corrective blend shape function makes predictions only about a subset of the mesh vertices. 
We adopt the same notation used in \smpl \cite{SMPL:2015}. We start with an artist defined template, $\overline{\vec{T}} \in \mathbb{R}^{3N}$ in the rest pose $\vec{\theta}^*$ (i.e. T-Pose) where  $N=6890$ is the number of mesh vertices. The model kinematic tree contains $K=24$ joints,  corresponding to $23$ body joints in addition to a root  joint. The template $\overline{\vec{T}}$ is then deformed by a shape corrective blend shape function $B_S$ that captures the subjects identity and a function $B_P$ that adds correctives offsets such that mesh looks realistic when posed.  
\paragraph{Shape Blend Shapes.} The shape blend shape function $B_S(\vec{\beta};\mathcal{S}):\mathbb{R}^{|\vec{\beta}|} \rightarrow \mathbb{R}^{3N}$ maps the identity  parameters $\vec{\beta}$ to vertex offsets from the template mesh as
\begin{equation}
B_S(\vec{\beta};\mathcal{S}) = \sum_{n=1}^{|\vec{\beta}|} \beta_{n} S_n,
\end{equation}
where $\vec{\beta} = [ \beta_{1}, \cdots , \beta_{|\beta|} ]$ are the shape coefficients, and $\mathcal{S} = [S_{1}, \cdots, S_{|\beta|}] \in \mathbb{R}^{3N\times|\vec{\beta}|}$ are the principal components capturing the space of human shape variability. The shape correctives are added to the template:
\begin{equation}
\vec{V}_{shaped} =  \overline{\vec{T}} +   B_S(\vec{\beta};\mathcal{S}),
\end{equation}
where $\vec{V}_{shaped}$ contains the vertices representing the subject's physical attributes and identity.

\paragraph{Pose and Shape Corrective Blend Shapes.}  
The output of the shape corrective blend shape function, $\vec{V}_{shaped}$,  is further deformed by a  pose corrective function.  The pose corrective function is conditioned on both pose and shape and adds corrective offsets such that, when the mesh is posed with LBS, it looks realistic. We denote the kinematic tree unit quaternion vector as $\vec{q} \in \mathbb{R}^{96}$ (24 joints each represented with 4 parameters). The pose corrective function is denoted as $B_P({\vec{q}},\beta_2) \in \mathbb{R}^{|\vec{q}| \times 1} \rightarrow \mathbb{R}^{3N}$, where $\beta_2$ is the PCA coefficient of the second principal component, which highly correlates with the body mass index (BMI) as shown in \supmat. The \modelname pose corrective function is factored into a sum of pose corrective functions: 
\begin{equation}
\label{eq:smplr_blends}
B_P({\vec{q}},\beta_2;\mathcal{K},\mathbf{A}) = \sum_{j=1}^{K-1} {B}^{j}_P({\vec{q}_{ne(j)}},\beta_2;{\mathcal{K}_j},{\vec{A}_j}),
\end{equation}
where a pose corrective function is defined for each joint in the kinematic tree excluding the root joint. 
The per-joint pose corrective function ${B}^{j}_P({\vec{q}_{ne(j)}},\beta_2;{\mathcal{K}_j},{\vec{A}_j})$ predicts corrective offsets given $\vec{q}_{ne(j)} \subset \vec{q}$, where $\vec{q}_{ne(j)}$ is a set containing the joint $j$ and its direct neighbors in the kinematic tree. 
This formulation results in more powerful regressors compared to \smpl.
$\mathcal{K}_j \in \mathbb{R}^{3N \times |\vec{q}_{ne(j)}| + 1}$ is a linear regressor weight matrix and $\vec{A}_j$ are the activation weights for each vertex, both of which are learned.
 Each pose corrective function,  $B^{j}_P(\vec{q}_{ne(j)},\beta_2)$ , is defined as a composition of two functions, an activation function and a pose corrective regressor.

\paragraph{Activation Function.} For each  joint, $j$, we define a learnable set of mesh vertex weights, ${\vec{A}_j} = [w_j^1, \cdots, w_j^N]  \in \mathbb{R}^{N}$, where $w^i_j \in \mathbb{R}$ denotes the weight of the $i^{\mathit th}$ mesh vertex with respect to the $j$  joint. 
The weight $w^i_j$ for each vertex $i$ is initialized as the reciprocal of the minimum geodesic to the set of vertices around joint $j$, normalized to the range $[0,1]$.
The weights are thresholded by a non-linear activation function, specifically a rectified linear unit (ReLU):
\begin{equation}
  \phi(w^i_j)=\begin{cases}
    0, & \text{if $w^i_j  \leq 0$},\\
    w^i_j , & \text{otherwise},
  \end{cases}
  \label{eq:activation}
\end{equation}
such that during training, vertices with a $w^i_j \leq 0 $ have weight $0$. The remaining set of vertices with  $w^i_j > 0$ defines the support region of joint $j$.   

\paragraph{Pose Corrective Regressor.}  
The per-joint pose corrective function is defined as $P_j(\vec{q}_{ne(j)}) \in \mathbb{R}^{|\vec{q}_{ne(j)}| + 1} \to \mathbb{R}^{3N}$, which regresses corrective offsets given the joint and its direct neighbors' quaternion values
\begin{equation}
 P_j(\vec{q}_{ne(j)},\beta_2;\mathcal{K}_j) =  \mathcal{K}_j((\vec{q}_{ne(j)} -\vec{q}_{ne(j)}^*)^T | \beta_2)^T,
 \label{eq:joint_corrective}
\end{equation}
where $\vec{q}_{ne(j)}^*$ is the vector of quaternion values for the set of joints $ne(j)$ in rest pose, and $\beta_2$ is concatenated to the quaternion difference vector.  $\mathcal{K}_j \in \mathbb{R}^{3N \times |\vec{q}_{ne(j)}| + 1}$ is the regression matrix for joint $j$'s pose correctives offsets. The predicted pose corrective offsets in Equation~\ref{eq:joint_corrective} are masked by the joint activation function:
\begin{equation}
\label{eq:local_blends}
{B}^{j}_P({\vec{q}_{ne(j)}};\vec{A}_j, \mathcal{K}_j)=  \phi({\vec{A}_j})   \circ  P_j(\vec{q}_{ne(j)},\beta_2;\mathcal{K}_j),
\end{equation}
where $ \vec{X}  \circ \vec{Y} $ is the element wise Hadamard product between the vectors $\vec{X}$ and $\vec{Y}$.
During training, vertices with zero activation with respect to joint $j$, will have no corrective offsets added to them. Therefore when summing the contribution of the individual joint pose corrective functions in Equation~\ref{eq:smplr_blends},  each joint only contributes pose correctives to the vertices for which there is support.

\paragraph{Blend Skinning.} Finally, the mesh with the added pose and shape corrective offsets is transformed using a standard skinning function $W(\overline{\vec{T}},\vec{J},\vec{\theta},\mathcal{W})$ around the joints,  $\vec{J} \in \mathbb{R}^{3K}$ and linearly smoothed by a learned set of blend weight parameters $\mathcal{W}$. 
The joint locations are intuitively influenced by the body shape and physical attributes. Similar to \smpl, the joints $\vec{J}(\vec{\beta};\mathcal{J},\overline{\vec{T}},\mathcal{S}) = \mathcal{J}(\vec{V}_{shaped})$ are regressed from $\vec{V}_{shaped}$ by a sparse function $\mathcal{J}: \mathbb{R}^{3N} \rightarrow \mathbb{R}^{3K}$. 

To summarize, \modelname is full defined by:
\begin{equation}
M(\vec{\beta},\vec{\theta})= W(T_p(\vec{\beta},\vec{\theta}),J(\vec{\beta}),\vec{\theta},\bf{\mathcal{W}}),
\end{equation}
where $T_P$ is defined as:
\begin{equation}
T_p(\vec{\beta},\vec{\theta}) = \overline{\vec{T}} + B_S(\vec{\beta}) + B_P(\vec{q},\beta_2),
\end{equation}
where $\vec{q}$ is the quaternion representation of pose $\vec{\theta}$.
The \modelname model is fully parameterized by $72$ (i.e. 24 * 3) pose parameters $\vec{\theta}$ in axis-angle representation, and up to 300 shape parameters $\vec{\beta}$.

\subsection{Model Training} 
\modelname{} training is similar to \smpl{}
\cite{Loper:SIGASIA:2014}. The key difference is the training of the pose corrective function in Equation~\ref{eq:smplr_blends}. \modelname{} pose corrective blend shapes are trained to minimize the \emph{vertex-to-vertex} error between the model predictions and the ground-truth registrations where, in each iteration, the model parameters ($\mathcal{A}$,$\mathcal{K}$) are minimized by stochastic gradient descent across a batch of B registrations, denoted as $\vec{R} \in \mathbb{R}^{3N}$. 
The data term is given by: 
\begin{equation}
\mathcal{L}_D = \frac{1}{B} \sum_{i=1}^B || M(\vec{\beta}_i,\vec{\theta}_i) - \vec{R}_i||_2.
\end{equation}
In addition to the data term we regularize the pose corrective regression weights ($\mathcal{K}$) with an $L2$ norm:
\begin{equation}
\mathcal{L}_{B}=  \lambda_b \sum_{i=1}^{K-1}||\mathcal{K}_i ||_{2} , 
\end{equation}
where $K$ is the number of joints in \modelname{} and  $\lambda_b$ is a scalar constant. In order to induce sparsity in the activation masks $\phi(.)$, we use an $L1$ penalty
\begin{equation}
\mathcal{L}_A=  \lambda_c ||\sum_{i=1}^{K-1} \phi_j(\vec{A}_j)||_1 , 
\label{eq:sparse_mask}
\end{equation}
where $\lambda_c$ is a scalar constant. Similar to \smpl{} we use a
sparsity regularizer term on the skinning weights $\mathcal{W}$ and regularize the skinning weights to initial artist-defined skinning weights, $\mathcal{W}_\text{prior} \in \mathbb{R}^{N\times K}$:
\begin{equation}
\mathcal{L}_W=  \lambda_p||\mathcal{W}-\mathcal{W}_\text{prior}||_2 + \lambda_s||\mathcal{W}||_1 , 
\end{equation}
where $\lambda_p$ and $\lambda_s$ are scalar constants. To summarize the complete training objective is given by 
\begin{equation}
\mathcal{L} = \mathcal{L}_D + \mathcal{L}_B + \mathcal{L}_A + \mathcal{L}_W.
\label{eq:full_obj}
\end{equation}
The objective in Equation~\ref{eq:full_obj} is minimized with respect to the skinning weights $\mathcal{W}$, pose corrective regression weights $\mathcal{K}_{1:24}$, activation weights $\vec{A}_{1:24}$. We train the model iteratively. In each training iteration, we anneal the regularization parameters as described in the Sup.~Mat.
\section{Experiments}
\label{sec:experiments}
\begin{figure}[t]
	\begin{subfigure}{.5\textwidth}
\centering 
\includegraphics[width=\textwidth]{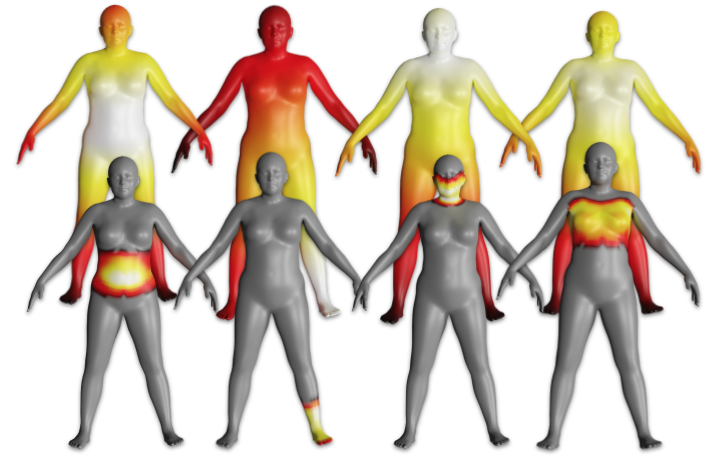}
\caption{}
\label{fig:activation}
	\end{subfigure}
	\begin{subfigure}{.5\textwidth}
	\centering 
	\includegraphics[width=1\textwidth]{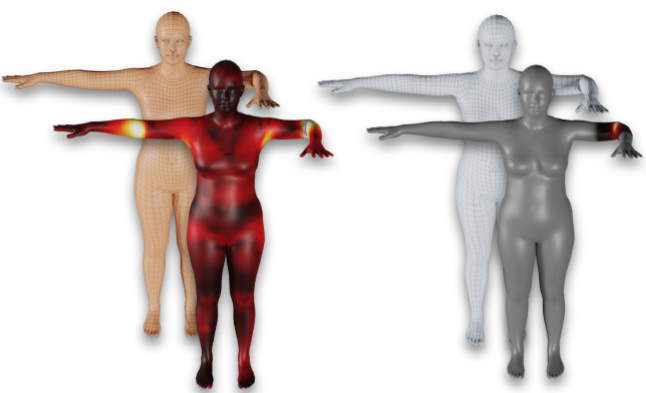}
	\caption{}
	\label{fig:activation_star}
\end{subfigure}
\caption{\textbf{Spatially local and sparse pose corrective
    blend shapes}. 
(a) The top row shows a sample of the joints activation functions output before training and the
bottom row shows the output after training (gray is zero).
(b) shows SMPL (brown) and STAR (white) in the rest pose except for the
left elbow, which is rotated.
The heat map visualizes the corrective offsets for each model
caused by moving this one joint.
Note that unlike STAR, SMPL has spurious long-range displacements.}
\end{figure}

\subsection{Activation} Key to learning the sparse and spatially local
pose corrective blend shapes are the joint activation functions
introduced in Equation \ref{eq:activation}. 
During training the output of the activation functions becomes more sparse, limiting the number of vertices a joint can influence. Figure \ref{fig:activation} summarizes a sample of the activation functions output before and after training.  As a result of the output of the activation functions becoming more sparse, the number of model parameters decreases. By the end of training, the male model pose blend shapes contained $3.37\times10^5$ non-zero parameters and the female model contained $3.94\times10^5$ non-zero parameters compared to \smpl which has a dense pose corrective blendshape formulation with $4.28\times10^6$ parameters. At test time only the  non-zero parameters need to be stored.

Figure \ref{fig:activation_star} show a SMPL model bending an elbow
resulting in a bulge in the other elbow, as a result of the pose
corrective blend shapes learning long range spurious correlations from
the training data. 
In contrast, \modelname correctives are spatially local and sparse, this is a result of the learned local sparse pose corrective blend shape formulation of \modelname. 

\subsection{Model Generalization}
\label{sec:model_generalization}
\begin{figure}[t]
	\includegraphics[width=\textwidth]{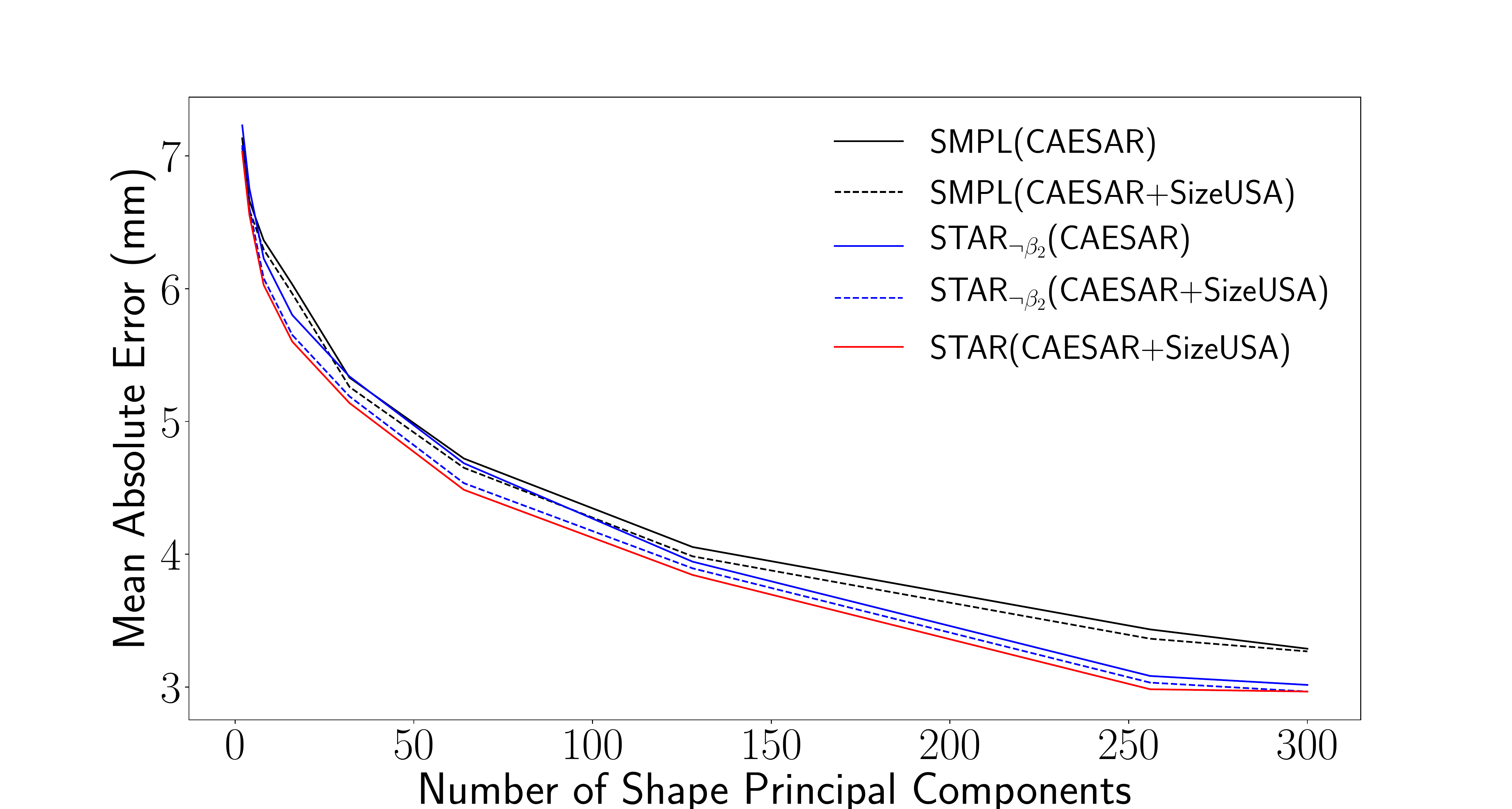}  
\caption{\textbf{Generalization Accuracy}: Evaluating \modelname and \smpl on unseen bodies. STAR$_{\lnot\beta_2}$(CAESAR) is \modelname trained on CAESAR with pose correctives depending on pose only (i.e. independent of $\beta_2$), STAR$_{\lnot\beta_2}$(CAESAR+SizeUSA) is \modelname trained on CAESAR and SizeUSA with pose corrective blend shapes depending on pose only, and STAR(CAEAR+SizeUSA) is STAR trained on CAEASAR and SizeUSA with pose and shape dependent pose corrective blend shapes.} \label{fig:4}
\end{figure}

\begin{figure}[t]
\includegraphics[width=0.90\textwidth]{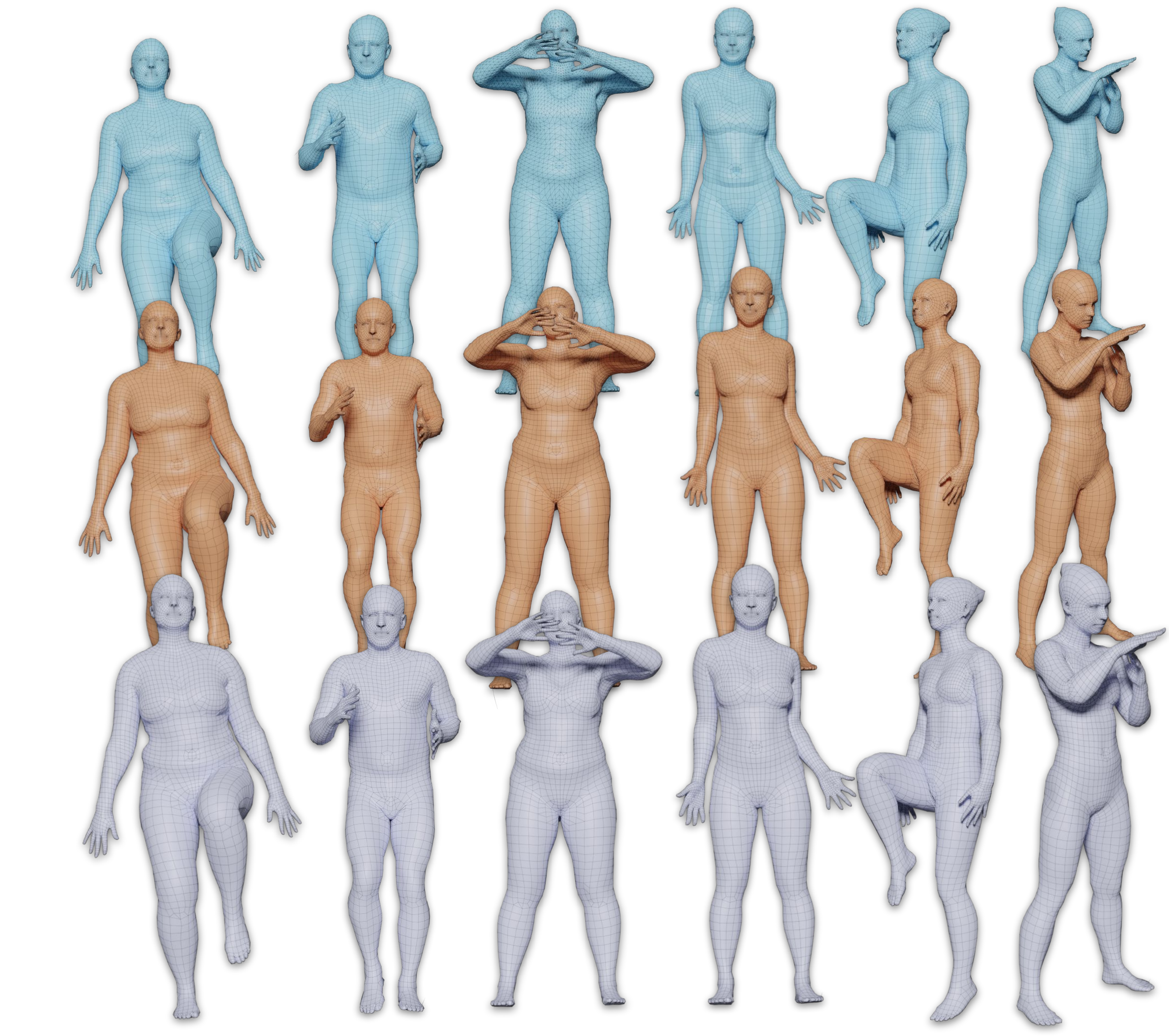}
\caption{\textbf{Qualitative Evaluation}:  Comparison between \smpl and \modelname. The ground truth registrations are shown in blue, the corresponding \smpl model fit meshes are shown in brown and \modelname fits are shown in white. Here, both \modelname and \smpl are trained on the CAESAR database.}
	\label{fig:5}
\end{figure}

While the learned activation masks are sparse and spatially local, which is good, it is equally important that the model still generalizes to unseen bodies. 
To this end, we evaluate the model generalization on held out test subjects. The test set we use contains the publicly available Dyna dataset \cite{DynaURL} (the same evaluation set used in evaluating the \smpl model), in addition to the 3DBodyTex dataset \cite{saint20183dbodytex} which contains static scans for 100 male and 100 female subjects in a diversity of poses. 
The total test set contains 570 registered meshes of 102 male subjects and 104  female subjects. 
We fit the models by minimizing the vertex to vertex mean absolute error (v2v), where the pose $\vec{\theta}$ and shape parameters $\vec{\beta}$ are the free optimization variables. 
We report the mean absolute error in (mm) as a function of the number of used shape coefficients in Figure~\ref{fig:4}. 
We first evaluate SMPL and \modelname when they are both trained using the CAESAR dataset. 
In this evaluation both models are trained on the exact same pose and shape data. 
Since they both share the same topology and kinematic tree, differences in the fitting results are solely due to the different formulation of the two models. 
In Figure \ref{fig:4}, \modelname uniformly generalizes better than SMPL on the unseen test subjects. 
A sample qualitative comparison between \smpl and \modelname fits is shown in Figure \ref{fig:5}.
 
\subsection{Extended Training Data}
\begin{figure*}[t]
\begin{subfigure}{.5\textwidth}
\includegraphics[width=\linewidth]{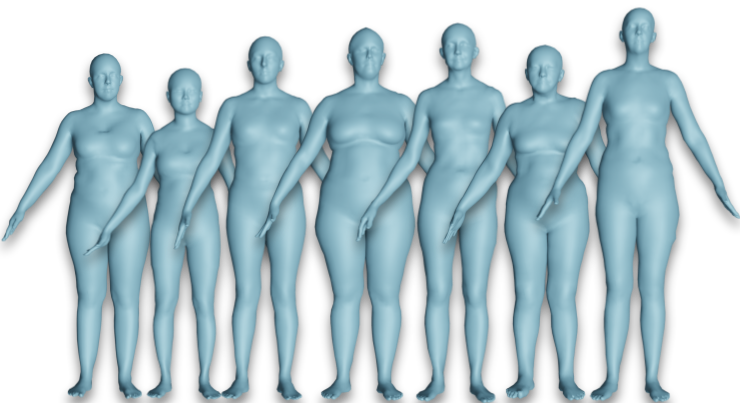}
\caption{Sample Females: CAESAR}
\label{fig:caesar}
\end{subfigure}
\begin{subfigure}{.5\textwidth}
\includegraphics[width=\textwidth]{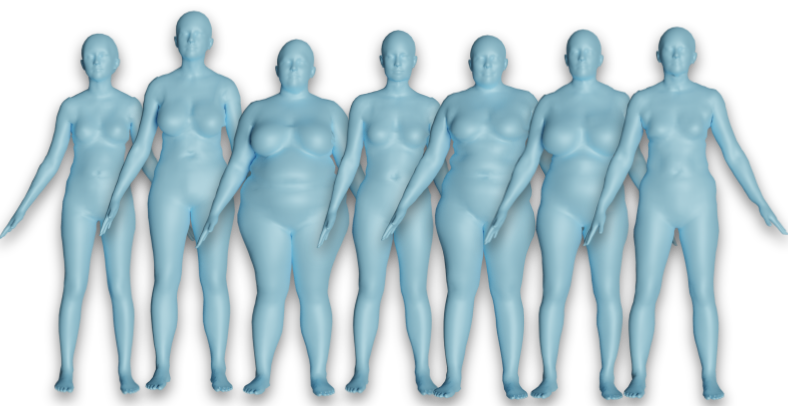}
\caption{Sample Females: SizeUSA}
\label{fig:sizeusa}
\end{subfigure} \newline 
\begin{subfigure}{.5\textwidth}
\includegraphics[width=\textwidth]{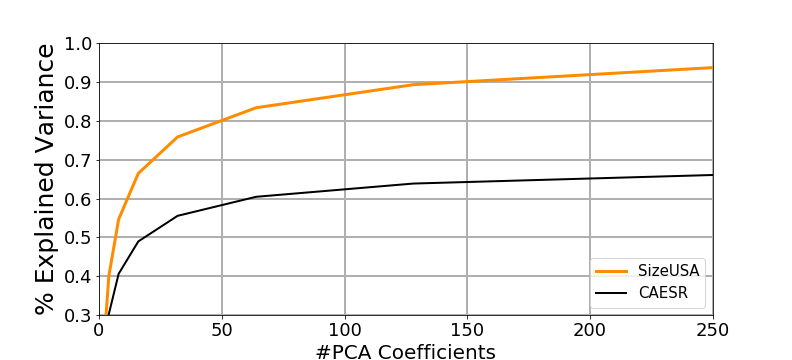}
\caption{Shape Space Trained on SizeUSA}
\label{fig:sizeusa_reconc}
\end{subfigure}
\begin{subfigure}{.5\textwidth}
\includegraphics[width=\textwidth]{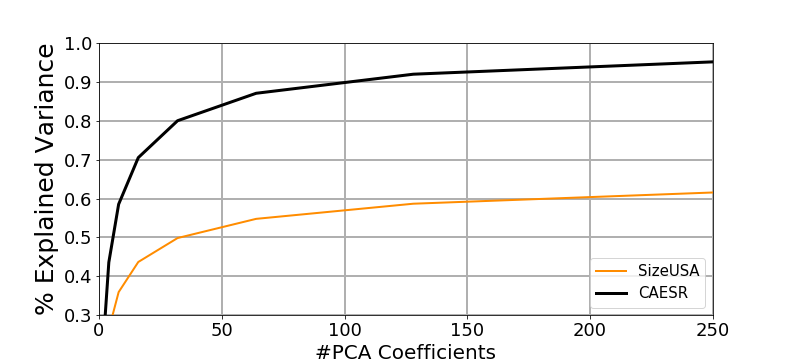}
\caption{Shape Space Trained on CAESAR}
\label{fig:caesar_reconv}
\end{subfigure}
\caption{\textbf{Explained Variance:} The percentage of explained variance of SizeUSA and CAESAR subjects when shape space is trained on SizeUSA is shown in Figure \ref{fig:sizeusa_reconc} and when the shape space is trained on CAESAR subjects in Figure \ref{fig:caesar_reconv}.}
\label{fig:sizeusa_redundancy}
\end{figure*}

The CAESAR dataset is limited in its diversity, consequently limiting model generalization.
Consequently, we extend the shape training database to include the SizeUSA database~\cite{SizeUSA}. 
SizeUSA contains low quality scans of 2845 male and 6434 females with ages  varying between 18 to 66+; a sample of the SizeUSA bodies compared to the CAESAR bodies are shown in Figure \ref{fig:caesar} and Figure \ref{fig:sizeusa}. 
We evaluate the generalization power of models trained separately on CEASER and SizeUSA.
We do so by computing the percentage of explained variance of the SizeUSA subjects given a shape space trained on the CAESAR subjects, and vice versa.
The results are shown in Figure \ref{fig:sizeusa_redundancy} for the female subjects, the full analysis for both male and female subjects is shown in the \supmat. 
The key insight from this experiment is that a shape space trained on a single data set was not sufficient to explain the variance in the other data set. 
This suggests that training on both dataset should improve the model shape space expressiveness. 

We retrain train both \modelname and \smpl  on the combined CAESAR and SizeUSA datasets an evaluate the model generalization on the held out test set as a function of the number of shape coefficient used as shown in Figure \ref{fig:4}. 
Training on both CAESAR and SizeUSA results in both \smpl and \modelname generalizing better than when trained only on CAESAR.
We further note  that \modelname still uniformly generalizes better than \smpl when both models are trained on the combined CAESAR and SizeUSA dataset. 
Importantly \modelname is more accurate than SMPL despite the fact that uses many fewer parameters.
Finally we extend the pose corrective blend shapes of \modelname to be conditioned on both body pose and body shape and evaluate the model on the held out set.
This results in a further improvement in the model generalization accuracy that, while modest, is consistent.

\section{Discussion}
\modelname has 93 pose corrective blend shapes compared to 207 in \smpl and is 80\% smaller than \smpl. 
It is surprising that it is able to uniformly perform better than \smpl when trained on the same data.
This highlights the fact that the local and sparse assumptions of the pose corrective blend shapes is indeed realistic a priori knowledge that should be incorporated in any body model. 
Importantly, having fewer parameters means that \modelname is less likely to overfit, even though our non-linear model makes training more difficult.

For \smpl, the authors report that enforcing sparsity of the pose corrective blend shapes resulted into worse results than \smpl. 
We take a different approach and learn the sparse set of vertices relevant to a joint from data. 
The key strength of our approach is that it is learned from data. 

We are able to learn spatially local and sparse joint support regions due to  two key implementation details: 
The initialization of the vertex weight $\vec{A}_j$ with the normalized inverse of geodesic distance to a joint. 
Secondly, the pose corrective blend shapes for each joint are regressed from local pose information, corresponding to the joint and its direct neighbors in the kinematic tree; this is a richer representation than SMPL.
These two factors together with the sparsity inducing $L1$ norm on the activation weights, act as an inductive bias to learn a sparse set of vertices that are geodesically local to a joint. 

The sparse pose correctives formulation reduces the number of parameters and regularizes the model, preventing it from learning spurious long range correlations from the training data.  
Since each vertex is only influenced by a limited number of joints in the kinematic tree, the gradients propagated through the model are sparse and the derivative of a vertex with respect to a geodesically distant joint is 0, which is not the case in the \smpl.     

\section{Conclusion}
\label{sec:conclusion}
We have introduced \modelname, which has fewer parameters than \smpl  yet is more accurate and generalizes better to unseen bodies when trained on the same data. 
Our key insight is that human pose deformation is local and sparse.  While this observation is not new, our formulation is. We define a non-linear (ReLU) activation function for each joint and train the model from data to estimate both the linear corrective pose blend shapes and the activation region on the mesh that these joints influence.
We kept what is popular with \smpl while improving on it in every sense. \modelname has only 20\% of the pose corrective parameters of \smpl. 
Our training method and localized model fixes a key problem of \smpl -- the spurious,  long-range, correlations that result in non-local deformations. Such artifacts make \smpl unappealing for animators. Moreover, we show that, while \smpl is trained from thousands of scans, human bodies are more varied than the CAESAR dataset. More training scans results in a better model.
Finally we make pose-corrective blend shapes depend on body shape, producing more realistic deformations.
We make \modelname available for research with 300 shape principal components.
It can be swapped in for \smpl in any existing application since the pose and shape parameterization is the same to the user.
Future work work should extend this approach to the SMPL-X model which includes an expressive face and hands.
\newline 
\textbf{Acknowledgments:} The authors thank N.~Mahmood for
insightful discussions and feedback, and the International Max Planck Research
School for Intelligent Systems (IMPRS-IS) for supporting
A.~A.~A.~Osman. The authors would like to thank Joachim Tesch, Muhammed Kocabas, Nikos Athanasiou, Nikos Kolotouros and Vassilis Choutas for their support and fruitful discussions.   \newline
\textbf{Disclosure:} In the last five years, MJB has received research gift funds from Intel, Nvidia, Facebook, and Amazon. He is a co-founder and investor in Meshcapade GmbH, which commercializes 3D body shape technology. While MJB is a part-time employee of Amazon, his research was performed solely at, and funded solely by, MPI.

\end{document}